# A Binary Schema and Computational Algorithms to Process Vowel-based Euphonic Conjunctions for Word Searches

**Kasmir Raja S. V.[1], Rajitha V.[2,3], Meenakshi Lakshmanan[2]**

[1] *Dean – Research, SRM University, Chennai 600 005, India*
[2] *Department of Computer Science, Meenakshi College for Women, Chennai 600 024, India*
[3] *Research Scholar, Mother Teresa Women's University, Kodaikanal 624 101, India*
*warftrajitha@gmail.com*

**Abstract**

Comprehensively searching for words in Sanskrit E-text is a non-trivial problem because words could change their forms in different contexts. One such context is sandhi or euphonic conjunctions, which cause a word to change owing to the presence of adjacent letters or words. The change wrought by these possible conjunctions can be so significant in Sanskrit that a simple search for the word in its given form alone can significantly reduce the success level of the search. This work presents a representational schema that represents letters in a binary format and reduces Pāṇinian rules of euphonic conjunctions to simple bit set-unset operations. The work presents an efficient algorithm to process vowel-based sandhis using this schema. It further presents another algorithm that uses the sandhi processor to generate the possible transformed word forms of a given word to use in a comprehensive word search.

## 1. Introduction
Over the past decade, a large and increasing number of philosophical, grammatical, literary and other works in Sanskrit have become available for reference as E-texts in digital libraries [15-18]. Many of the old printed books that are the sources of the Sanskrit E-texts do not have extensive tables of contents or detailed word-indices. As such, a cumbrous manual search is usually needed to locate specific words or phrases of interest in them. With the E-texts in Unicode, readers have the great advantage of being able to search rapidly and with ease for what they seek.

Comprehensively searching for and effectively locating specific Sanskrit words or phrases is important and useful in a variety of ways to modern researchers as well as to traditional Pundits. For instance, the presence or absence of certain phrases and the frequency of occurrence of specific words have been proposed and used as factors to investigate the authorship of several philosophical works. In a couple of instances, certitude about the position of the author on an issue of importance has been arrived at by examining whether anywhere in his work he has used two key words synonymously. Pundits are normally in a position to recall many words and phrases of the traditional, authoritative texts that they expound and often specify portions of texts by means of key words and phrases. Several scholarly debates and books have focussed on where and how some profound words have been used in scriptural texts. In view of the availability for reference of a large repertoire of Sanskrit E-texts and the aforesaid importance and usefulness of automated word-searches to researchers as well as to traditional Pundits, a computational algorithm is required for comprehensively and effectively locating in the Sanskrit E-texts, specified words and phrases. The reason for directing attention to Sanskrit E-texts is the extent of the complexity involved in *comprehensively* searching for Sanskrit words and phrases; it is demonstrably insufficient to search for just the specified word as is done in the case of English and as is generally being done at present even in the case of Sanskrit E-texts.

## 2. The Problem

Unlike in the case of English E-texts, to search comprehensively for a word or phrase in a Sanskrit E-text, euphonic conjunctions ('*sandhis*') need to be considered. A *sandhi* is a point in a word or between words, at which adjacent letters coalesce and transform. This is a common feature of Indian languages and is particularly elaborately dealt with and used in Sanskrit. For example, the compound word '*asamardhiḥ*' (meaning unmatched affluence) can, owing to there being options with regard to a euphonic conjunction within it, also be written as '*asamarddhiḥ*'. Moreover, were it to be preceded in a sentence by a word ending with '*a*' and followed by a word beginning with '*a*', it could even be encountered as '*āsamarddhir*'. Decidedly, a simple search for '*asamardhiḥ*' would not yield '*āsamarddhir*', the equivalent form in which it may be present in the document. Existing search engines [19] do not provide such comprehensive search capabilities.

Thus, all the grammatically permissible euphonic conjunctions that are pertinent to the Sanskrit word or phrase of interest ought to be considered, the various resulting forms in which the word or phrase may be encountered in the text generated and all of them duly searched for.

Since searching is itself a costly operation if the given source text is large, which is often the case with ancient Sanskrit works, the *sandhi* processing needs to be done extremely quickly in spite of the complexity and number of *sandhi* rules required to be processed. Thus, the problem includes the performance parameter as well.

### 2.1. Language Representation

The Unicode (UTF-8) standard is what has been adopted universally for the purpose of encoding Indian language texts into digital format. The Unicode Consortium has assigned the Unicode hexadecimal range 0900 - 097F for Sanskrit characters, i.e. the *Devanāgarī* script.

All characters including the diacritical characters used to represent Sanskrit letters in E-texts are found dispersed across the Basic Latin (0000-007F), Latin-1 Supplement (0080-00FF), Latin Extended-A (0100-017F) and Latin Extended Additional (1E00 – 1EFF) Unicode ranges.

The Latin character set has been employed in this work to represent Sanskrit letters as E-text. The schema and algorithms presented do not use *Devanāgarī* script. To use the algorithms for text that is in *Devanāgarī* script, the text needs to be converted to Latin text that uses diacritical marks for the special letters in Sanskrit.

### 2.2. The Basis of this Work

While working with or processing Sanskrit grammar, authenticity is acknowledged only if the work is based on the rules of Sanskrit grammar codified by the ancient grammarian Pāṇini in his celebrated work, 'Aṣṭādhyāyī' (which simply means 'work in eight chapters'). The work contains comprehensive information codified in the form of terse *sūtras* or aphorisms and is saluted universally as the ultimate authority on Sanskrit grammar. The work is also hailed as the product of an extraordinarily sharp mind, for mastering it requires years of study along with commentaries such as the *Mahābhāṣya* of Sage Patañjali, the *Siddhānta-kaumudī* of Bhaṭṭoji Dīkṣita and the *Kāśikā* of Vāmana and Jayāditya.

For the current work that deals with the processing of euphonic conjunctions, the *Siddhānta-kaumudī* [1] and the *Kāśikā* [2] have been used as the bases. Both are recognized by scholars as authentic and unabridged commentaries on Pāṇini's work. The former has a different ordering of sutras in a convenient topic-wise categorization, while the latter preserves the crucial ordering inherent in the original work of Pāṇini.

### 2.3. The *Māheśvara-sūtras*

The *Māheśvara-sūtras*, or the 'aphorisms of Maheśvara', are said to have emanated from a special drum called 'ḍamaru' (hourglass drum) held in the hand of Lord Maheśvara (a form of God in the Hindu pantheon). These sounds constitute the entire Sanskrit alphabet ordered in a specific sequence. The aphorisms of Pāṇini are framed using these *Māheśvara* aphorisms as basis, and cannot be understood without them. There are fourteen *Māheśvara* aphorisms and these are listed below:

1. *a-i-u-ṇ*
2. *ṛ-ḷ-k*
3. *e-o-ṅ*
4. *ai-au-c*

   5. *ha-ya-va-ra-ṭ*
   6. *la-ṇ*
   7. *ña-ma-ṅa-ṇa-na-m*
   8. *jha-bha-ñ*
   9. *gha-ḍha-dha-ṣ*
   10. *ja-ba-ga-ḍa-da-ś*
   11. *kha-pha-cha-ṭha-tha-ca-ṭa-ta-v*
   12. *ka-pa-y*
   13. *śa-ṣa-sa-r*
   14. *ha-l*

The last letter in each of the above aphorisms is only a place-holder and is not counted as an actual letter of the aphorism. The first four aphorisms list the short forms of all the vowels, while the rest list the consonants. It must be noted that the letter '*a*' added to each of the consonants is only to facilitate pronunciation and is not part of the consonant proper.

## 3. The Approach

There are more than seventy aphorisms of Pāṇini that deal with *sandhis*. These aphorisms lay out the rules for *sandhi* transformations, giving the conditions under which they take place.

The challenge is to develop a computational algorithm to handle the entire range of *sandhis*. Such a computational algorithm would be useful to search for various forms of a given word in Sanskrit, thus enabling a thorough search of Unicode texts. The task of arriving at the resultant transformed word given two words may be an easy one for a human being thoroughly conversant with the grammatical rules of Sanskrit, but it is a computationally non-trivial task, given the complexity of the rules. Existing methods of *sandhi* processing are based on a derived understanding of the functioning of euphonic conjunctions and do not start from the *Pāṇini-sūtras*. Further, finite automata, HMM or artificial intelligence methods [3-13] have been used in them. There has been one work that has formulated a mathematical model of the original aphorisms that lends itself to speedy processing [14].

The present work too directly codifies Pāṇini's rules in a novel way using binary representations. A unique data representation has been devised in this work that simplifies rules to minimal binary set-unset operations. As such, this work presents an enhanced-efficiency alternative to the earlier methods. Furthermore, it simplifies the understanding of the entire grammar related to euphonic conjunctions.

## 4. The Binary Schema

A point of *sandhi* is denoted by

$$x + y$$

where $x$ and $y$ denote the *sandhi* letters and the symbol '+' denotes adjacency. The variable $X$ denotes the sequence of letters culminating in $x$; the variable $Y$ denotes the sequence of letters starting with $y$. The notations $X$ and $Y$ are used to depict special conditions that pertain to an entire word or sequence of letters involved in the *sandhi* rule.

*Sandhi* rules often involve the letter immediately preceding $x$ and/or the letter immediately succeeding $y$. Hence, we denote these letters as $u$ and $w$ respectively. As such, in cases where letters adjacent to the actual *sandhi* letters play a role in determining the outcome of the *sandhi*, we have

$$u + x + y + w$$

A unique schematic has been developed in this work to represent the letters of the Sanskrit alphabet. Any letter of the alphabet is represented in two parts. Table 1 gives the scheme used in this representation.

**Table 1: Binary representation scheme**

| # | Letters |
|---|---|
| 0 | a,ā,i,ī,u,ū,ṛ,ṝ,ḷ,e,ai,o,au |
| 1 | y,r,l,v,yaṁ,vaṁ,laṁ |
| 2 | k,kh,g,gh,ṅ,c,ch,j,jh,ñ,ṭ,ṭh,ḍ,ḍh,ṇ,t,th,d,dh,n,p,ph,b,bh,m,ś,ṣ,s,h |
| 3 | ṁ,ḥ,f,x,',# |
| 4 | a |
| 5 | a,ā |
| 6 | i,u,a |
| 7 | ī,ū,ā |
| 8 | i,u,[ṛ ḷ],a |
| 9 | ī,ū,ṝ,ā |
| 10 | i,u,ṛ,ḷ |
| 11 | ī,ū,ṝ |
| 12 | ṛ,ṝ,ḷ |
| 13 | e,o |
| 14 | ai,au |
| 15 | o,au,e,ai |
| 16 | e,o,ar,al |
| 17 | ār,ā̄r,āl |
| 18 | av,āv,ay,āy |
| 19 | ava |
| 20 | y,v,r,l |
| 21 | yaṁ,vaṁ,r,laṁ |
| 22 | [k kh g gh ṅ],[c ch j jh ñ],[ṭ ṭh ḍ ḍh ṇ],[t th d dh n],[p ph b bh m] |
| 23 | ś,ṣ,s |
| 24 | s |
| 25 | h |
| 26 | [k kh g gh],[c ch j jh],[ṭ ṭh ḍ ḍh],[t th d dh],[p ph b bh] |

| #  | Letters |
|----|---------|
| 27 | ṅ,ñ,ṇ,n,m |
| 28 | m |
| 29 | k,kh,g,gh,ṅ |
| 30 | c,ch,j,jh,ñ,ś |
| 31 | ṭ,ṭh,ḍ,ḍh,ṇ,ṣ |
| 32 | t,th,d,dh,n,s |
| 33 | p,ph,b,bh,m |
| 34 | k,kh,c,ch,ṭ,ṭh,t,th,p,ph,ś,ṣ,s |
| 35 | g,gh,ṅ,j,jh,ñ,ḍ,ḍh,ṇ,d,dh,n,b,bh,m,h |
| 36 | k,c,ṭ,t,p |
| 37 | kh,ch,ṭh,th,ph |
| 38 | g,j,ḍ,d,b |
| 39 | gh,jh,ḍh,dh,bh |
| 40 | g,gh,j,jh,ḍ,ḍh,d,dh,b,bh |
| 41 | k,kh,c,ch,ṭ,ṭh,t,th,p,ph |
| 42 | ṁ |
| 43 | ḥ |
| 44 | f |
| 45 | x |
| 46 | ' |
| 47 | # |

Part 1 denotes the category (serial number in Table 1) to which a letter belongs, and Part 2 denotes the term number within the series that the letter is or fits into. In any letter representation, Part 1 is a binary string of fixed length 48, in which the set bit denotes the category number, while Part 2 is a binary string of variable length, the maximum length being 16.

The first four (shaded) rows of the table stand for overall category rows, viz. vowels, semi-vowels, consonants and special characters respectively. One of these four bits have to be set in any letter representation. There is no corresponding Part 2 value for the bits 0, 1, 2 and 3 of Part 1.

Clearly, one letter has many representations in this scheme. For example, the letter '*e*' occurs in categories 13, 15 and 16, and is a vowel. Hence, the letter '*e*' has the following three representations:

**Representation 1**:
Part 1:

| Bit # | 0 | 1 | 2 | 3 | 4 | ... | 13 | 14 | 15 | 16 | 17 | 18 | ... | 46 | 47 |
|-------|---|---|---|---|---|-----|----|----|----|----|----|----|-----|----|----|
| Bit value | 1 | 0 | 0 | 0 | 0 | ... | 1 | 0 | 0 | 0 | 0 | 0 | ... | 0 | 0 |

Part 2:

| Bit # | 0 | 1 | 2 | 3 | 4 | … | 14 | 15 |
|-------|---|---|---|---|---|---|----|----|
| Bit value | 1 | 0 | 0 | 0 | 0 | … | 0 | 0 |

**Representation 2:**
Part 1:

| Bit # | 0 | 1 | 2 | 3 | 4 | ... | 13 | 14 | 15 | 16 | 17 | 18 | ... | 46 | 47 |
|---|---|---|---|---|---|---|---|---|---|---|---|---|---|---|---|
| Bit value | 1 | 0 | 0 | 0 | 0 | ... | 0 | 0 | 1 | 0 | 0 | 0 | ... | 0 | 0 |

Part 2:

| Bit # | 0 | 1 | 2 | 3 | 4 | ... | 14 | 15 |
|---|---|---|---|---|---|---|---|---|
| Bit value | 0 | 0 | 1 | 0 | 0 | ... | 0 | 0 |

**Representation 3:**
Part 1:

| Bit # | 0 | 1 | 2 | 3 | 4 | ... | 13 | 14 | 15 | 16 | 17 | 18 | ... | 47 | 48 |
|---|---|---|---|---|---|---|---|---|---|---|---|---|---|---|---|
| Bit value | 1 | 0 | 0 | 0 | 0 | ... | 0 | 0 | 0 | 1 | 0 | 0 | ... | 0 | 0 |

Part 2:

| Bit # | 0 | 1 | 2 | 3 | 4 | ... | 14 | 15 |
|---|---|---|---|---|---|---|---|---|
| Bit value | 1 | 0 | 0 | 0 | 0 | ... | 0 | 0 |

Thus the vowel *'e'* is represented in three ways as:
1. 1000 00000 00001 00000 00000 00100 00000 00000 00000 00000 | 10000 00000 00000 0
2. 1000 00000 00000 01000 00000 00100 00000 00000 00000 00000 | 00100 00000 00000 0
3. 1000 00000 00000 00100 00000 00100 00000 00000 00000 00000 | 10000 00000 00000 0

where the symbol '|' depicts the boundary between the two parts and the space between sets of five bits has been used in the depiction only for the sake of clarity.

As can be gleaned from the scheme in Table 1, for Part 1 of any letter, bit 0 will be set if and only if exactly one bit from bits 4 to 19 is set, bit 1 if and only if exactly one of bits 20 and 21 is set, bit 2 if and only if exactly one of bits 22 to 41 is set, and bit 3 if and only if exactly one of bits 42 to 47 is set. Further, not more than one bit out of bits 0, 1, 2 and 3 can be set for any letter. Though most of the letters found in categories 17 to 19 are not vowels proper, they are included because vowel *sandhi* rules use them. A bit sequence with all 0s in Part 1, indicates a null letter.

The sequel contains *sandhi* rules codified under this scheme. It would be difficult to denote every letter in the rules as a series of 0s and 1s, and would be cumbersome to read and decipher. Hence the following notation has been adopted to represent rules: $x_i(n) = 1$ indicates that the $n$th bit of Part $i$ of the letter $x$ is set, where $i = 1, 2$.

For example, $x_1(27) = 1$ means that bit 27 in Part 1 of $x$ is set to 1. Thus $x$ may be any of the five letters in category 27 of Table 1. For $x$ to denote the specific letter 'ñ', we would further have to specify, $x_2(1) = 1$; this would set bit 1 of Part 2 in $x$.

In Table 1, some categories have letters in sublists enclosed within [ ]. Only one bit of Part 2 is reserved for each such group in the list, and the bit for the sublist is set for any letter that is included in the sublist.

## 5.     Formation of Rules using the Schema

The above schema has been developed in order to simplify the rule processing.
In the delineation of rules and in the presentation of the algorithms, the key to symbols is as follows:
- // means single-line explanatory comment
- { } are block or set indicators
- ∧ denotes $and$
- ∨ denotes $or$
- ¬ denotes $not$
- ⊕ denotes $xor$
- | denotes word concatenation

Following is an example to show how the rule simplification has been achieved through this schema.

Consider the *sandhi* aphorism 8.4.40 of Pāṇini, "*stoḥ ścunāḥ ścuḥ //*". This aphorism means that if any dental (*t, th, d, dh, n*) or letter *s* is adjacent to a palatal (*c, ch, j, jh, ñ*) or letter *ś*, then the dental or *s* is replaced by the corresponding palatal or *ś* respectively. The aphorism 8.4.44, "*śāt //*", debars this rule from applying in the case of the specific palatal *ś* followed by a dental. The following examples illustrate how these rules are applied in practice:

>    *sa**t** + **c**it = sa**c** + cit*
>    *śārṅgi**n** + **j**aya = śārṅgi**ñ** + jaya*
>    *rāma**s** + **c**inoti = rāma**ś** + cinoti*
>    *pra**ś** + **n**aḥ = pra**ś** + **n**aḥ*     (no change)

The newly developed schema in this work would reduce the coding of these two rules to the following:

>    $if\ (x_1(30) \wedge \neg x_2(5)) \wedge y_1(32)$
>    {
>       $y_1(32) =\ 0$; //bit unset in Part 1
>       $y_1(30) =\ 1$; //bit set in Part 1
>    }
>    $elseif\ x_1(32) \wedge y_1(30)$
>    {
>       $x_1(32) =\ 0$; //bit unset in Part 1
>       $x_1(30) =\ 1$; //bit set in Part 1
>    }

Since the change is to the 'corresponding' palatal, and the letters in categories 31 and 33 have been arranged appropriately in the schema, the rule realization requires only a simple bit set and unset in Part 1 and no change in Part 2. Further, the $if$ conditions in the rule require only simple checks to see if the $x$ and $y$ belong to particular categories, i.e. if a particular bit is set in Part 1 for each of $x$ and $y$.

## 6. The Vowel-Sandhi Processor

The following is the algorithm to process the main rules pertaining to vowel *sandhis*. It has been codified as per the above schema. Further, when Part 1 or Part 2 of a letter is altered by a bit set operation, it is assumed that all set bits in that part are first unset before the operation.

**Algorithm VowelSandhiProcessor (*X, Y*)**
{

   //1. *avaṅ sphoṭāyanasya* || 6.1.123 ||
   //Common name: *avaṅādeśa sandhi*
   //If the word *go* is followed by a vowel
   //then the *o* of *go* is replaced by *ava*.
   if $X =$ '*go*' $\wedge\ y_1(0)$
   {
     $x_1(19) = 1;$
   } // further rules are processed

   //2. *eṅaḥ padāntādati* || 6.1.109 ||
   //Common name: *pūrvarūpa sandhi*
   //If *e* or *o* at the end of a word is followed by *a*,
   //then *e* or *o* remains, and the *avagraha* (ʼ) replaces *a*.
   if $x_1(13) \wedge\ y_1(4)$
   {
     $y_1(46) = y_1(3) = 1;$
     return $X|Y;$
   }

   //3. *akaḥ savarṇe dīrghaḥ* || 6.1.101 ||
   //Common name: *savarṇadīrgha sandhi*
   //If one of *a, i, u, ṛ* or *ḷ* or their long
   //equivalents *ā, ī, ū* and *ṝ* is followed by the
   //short or long form of the same letter,
   //then the corresponding long letter replaces both.
   if $(x_1(8) \vee x_1(9)) \wedge\ (y_1(8) \vee y_1(9)) \wedge \neg(x_2 \oplus y_2)$
   {
     delete $y;$
     $x_1(9) = 1;$

```
      return X|Y;
   }

   //4. omāṅośca || 6.1.95 ||
   //Common name: pararūpa sandhi
   //If a or ā is followed by o of the word om or oṁ,
   //then o replaces both.
   if x₁(5) ∧ ((y₁(13) ∧ y₂(1)) ∧ (w₁(28) ∨ w₁(42)))
   {
      delete x;
      return X|Y;
   }

   //5. etyedhatyūṭhsu || 6.1.89 ||
   //Common name: vṛddhi sandhi
   //Note: This rule clashes with rule 6.1.94 and Rule 6.1.87 and takes precedence.
   //For this rule, in all cases the resultant letter replaces x and y.
   //i) If a or ā is followed by eti or edhati, then vṛddhi letter ai replaces both
   //ii) If the preposition pra is followed by eṣa or eṣy, then vṛddhi letter ai replaces
   //both
   //iii) If a or ā is followed by ūh, then vṛddhi letter au replaces both
   //iv) If preposition pra is followed by ūḍh, then vṛddhi letter au replaces both
   //v) If word sva is followed by īr, then vṛddhi letter ai replaces both
   if x₁(5)  //x is 'a' or 'ā'
   {
      if (y₁(13) ∧ y₂(0))  //y is 'e'
      {
         if Y starts with {et, edhat}  //rule (i)
         {
            delete x;
            y₁(14) = 1;
            return X|Y;
         }
         elseif X = 'pra' ∧ Y starts with {eṣ, eṣy}  //rule (ii)
         {
            delete x;
            y₁(14) = 1;
            return X|Y;
         }
      }
      elseif (y₁(7) ∧ y₂(1))  //y is 'ū'
      {
```

```
            if w₁(25) //rule (iii)
            {
               delete x;
               y₁(14) = 1;
               return X|Y;
            }
            elseif X = 'pra' ∧ Y starts with {ūḍh} //rule (iv)
            {
               delete x;
               y₁(14) = 1;
               return X|Y;
            }
         }
         elseif X = 'sva' ∧ (y₁(11) ∧ y₂(0)) ∧ (w₁(20) ∧ w₂(2)) //rule (v)
         {
            delete x;
            y₁(14) = 1;
            return X|Y;
         }
      }

      //6. eṅi pararūpaṁ || 6.1.94 ||
      //Common name: pararūpa sandhi
      //If a or ā at the end of a preposition is followed by e or o,
      //then the e or o replaces both.
      //Note: The prepositions that qualify are: pra, ava, apa, upa, parā.
      if X ∈ {pra, ava, apa, upa, parā} ∧ y₁(13)
      {
         delete x;
         return X|Y;
      }

      //7. upasargādṛti dhātau || 6.1.91 ||
      //Common name: vṛddhi sandhi
      //i) If a or ā at the end of a preposition is followed by ṛ, ṝ or ḷ,
      //then vṛddhi letter ār, ār or āl respectively replaces both.
      //The prepositions that qualify are: pra, parā, apa, ava, upa
      //ii) If the word vatsara, kambala, vasana, daśa, ṛṇa is followed by the word ṛṇa,
      //then vṛddhi letter ār replaces both.
      //Note: This rule clashes with 6.1.87 (guṇa sandhi), and takes precedence.
      if X ∈ {pra, ava, apa, upa, parā} ∧ y₁(12)
      {
```

```
      delete x;
      y₁(17) =  1;
      return X|Y;
   }
   if X ∈ {vatsara, kambala, vasana, daśa, ṛṇa} ∧ Y = 'ṛṇa'
   {
      delete x;
      y₁(17) =  1;
      return X|Y;
   }
```

```
   //8. vṛddhireci || 6.1.88 ||
   //Common name: vṛddhi sandhi
   //If a or ā is followed by e, o, ai or au,
   //then the corresponding vṛddhi letter ai or au
   //replaces both.
   if x₁(5) ∧ ((y₁(13) ∨ y₁(14))
   {
      delete x;
      y₁(14) =  1;
      return X|Y;
   }
```

```
   //9. ādguṇaḥ || 6.1.87 ||
   //    uraṇ raparaḥ || 1.1.51 ||
   //Common name: guṇa sandhi
   //If a or ā is followed by i, ī, u, ū, ṛ, ṝ or ḷ,
   //then the corresponding guṇa letter e, o, ar or al
   //replaces both.
   if x₁(5) ∧ ((y₁(10) ∨ y₁(11))
   {
      delete x;
      y₁(16) =  1;
      return X|Y;
   }
```

```
   //10. ecoyavāyāvaḥ || 6.1.78 ||
   //Common name: ayāyāvāvādeśa sandhi
   //If e, o, ai or au is followed by a vowel,
   //then ay, av, āy, āv replace the first respectively.
   if x₁(15) ∧ y₁(0)
```

```
   {
      x₁(18) =  1;
      return X|Y;
   }

   //11. iko yaṇaci ‖ 6.1.77 ‖
   //Common name: yaṇādeśa sandhi
   //If i, ī, u, ū, ṛ, ṝ or ḷ is followed by a vowel,
   //then the corresponding semi-vowel (y, v, r, l)
   //replaces the first.
   if (x₁(10) ∨ x₁(11)) ∧ y₁(0)
   {
      x₁(20) =  x₁(1) = 1;
      return X|Y;
   }
}
```

## 7. The Search Engine

Algorithm VowelSandhiProcessor can be used to generate alternative forms of a given search word, if the word starts and/or ends in a vowel. The following algorithm is used to effect the generation of all the possible alternative forms of the given word, which are then searched for in the given text.

**Algorithm GenerateWordForms (*Z*)**

```
{
//Z is the search word.
//{WordForms} denotes the set of word forms generated by the
//algorithm and is initially the null set.
   Add Z to {WordForms};
   X = Z;
   for (each y in {vowels}, y = 'o' where Y in {om, oṁ})
   {
      Add VowelSandhiProcessor(X, Y) to {WordForms};
   }

   for (each Y ∈ {WordForms})
   {
      for (each X in {go, pra, ava, apa, upa, parā}, x in {vowels})
      {
         Add VowelSandhiProcessor(X, Y) to {WordForms};
      }
   }
}
```

## 8. Conclusion

The schema developed in this work is revolutionary on multiple counts. Firstly, the letter representation scheme is binary, and hence lends itself to speedy processing and also for implementation in an FPGA format. Secondly, the check involved in order to apply a Pāṇinian rule has been reduced to merely a check to see whether a bit is set or not. Thirdly, the transformation of a letter into another according to a rule is effected by a maximum of one bit-unset and another bit-set. Fourthly, the efficiency achieved by the above division of a letter representation into two parts and effecting change-of-letter operations by simply changing the first part through minimal bit operations is unprecedented in the literature.

The representation schema and the results of the *sandhi*-processing algorithm represent an efficient computational model to process the vowel-based euphonic conjunctions in Sanskrit.

The proof of the efficiency of the *sandhi* processing algorithm and the optimality of the representational schema lies directly in the simplicity of the rule presentation that the schema facilitates.

The second algorithm presented in this work, which uses this *sandhi* processor to effect comprehensive word searches in texts is the first of its kind in the literature with regard to Sanskrit. The use of the *sandhi* processor for searching ensures comprehensiveness of the search, while the efficiency of the *sandhi* processing method presented in this work ensures that search speeds are not compromised due to the increase in the number of words to be searched for.

The schema devised in this work can be used as explained, to codify all consonant-based *sandhi* rules and other special *sandhi* rules of Pāṇini as well. Future work would include codification of all the non-vowel *sandhi* rules defined by Pāṇini, and expanding the conjoined word generation algorithm to include these *sandhis* as well.